\documentclass[conference]{IEEEtran}
\usepackage{stmaryrd}
\usepackage{amsfonts}
\usepackage{algorithmic}
\usepackage[ruled,vlined]{algorithm2e}

\usepackage{graphicx,times,amsmath} 

\IEEEoverridecommandlockouts    

\begin{document}

\title{\ \\ \LARGE\bf Ant Colony Optimization and Hypergraph Covering Problems\thanks{Ankit Pat is with the Cheriton School of Computer Science, University of Waterloo, Waterloo, Ontario, Canada (email: ankit.pat@uwaterloo.ca).}}

\author{Ankit Pat}


\maketitle

\begin{abstract}
Ant Colony Optimization (ACO) is a very popular metaheuristic for solving computationally hard combinatorial optimization problems. Runtime analysis of ACO with respect to various pseudo-boolean functions and different graph based combinatorial optimization problems has been taken up in recent years. In this paper, we investigate the runtime behavior of an MMAS*(Max-Min Ant System) ACO algorithm on some well known hypergraph covering problems that are NP-Hard. In particular, we have addressed the Minimum Edge Cover problem, the Minimum Vertex Cover problem and the Maximum Weak-Independent Set problem. The influence of pheromone values and heuristic information on the running time is analysed. The results indicate that the heuristic information has greater impact towards improving the expected optimization time as compared to pheromone values. For certain instances of hypergraphs, we show that the MMAS* algorithm gives a constant order expected optimization time when the dominance of heuristic information is suitably increased.
\end{abstract}


\section{Introduction}
\PARstart{A}{nt colony Optimization} (ACO), first proposed by Dorigo [1], is a class of nature-inspired stochastic algorithm which is widely used for various combinatorial optimization problems. The ACO algorithm is derived from the food searching behaviour of real ant colonies, who have demonstrated their ability to find the shortest path from their nest to the food source by stigmergic communication via pheromones. For a given problem, the ACO algorithm constructs a solution by random walk on a construction graph, with the walk influenced by pheromone values and heuristic information. Thus, ACO is able to incorporate problem specific knowledge in terms of heuristic information which separates it from other classes of Evolutionary Algorithms (EAs), etc.  Although ACO is not evolutionary in nature, it has by and large attracted the attention of Evolutionary Computing researchers because it is also a random search heuristic and both have overlapping application domains. Many different variants of the original ACO algorithm have been proposed and applied successfully to many real-world problems [2]. 
\par
Initial theoretical studies on ACO focussed on convergence properties and speed of different variants of ACO. St\"{u}tzle and Dorigo presented the convergence proof for the Max-Min ant system (MMAS [5]) in [4], while Gutjahr showed it for the Graph based ant system (GABS) [6]. The foundations of runtime analysis of ACO were laid in [8][9]. The analysis is typically done in a fashion that has been widely used for randomized algorithms and for simple evolutionary algorithms like (1+1) EA. In this approach, either the expected optimization time and/or the success probability after a certain number of steps are analysed. 
\par
Investigations on the runtime performance of ACO first started for various pseudo-boolean functions [3,10,11,12] where a simple ACO variant 1-ANT was analysed like (1+1) EA [7]. Another ACO variant MMAS has also been widely investigated for which the phase transition present in 1-ANT due to the evaporation factor does not occur [13,14]. The above techniques have been extended to different polynomial time combinatorial optimization problems, namely the Minimum Spanning Tree problem[15], Single Destination Shortest Path problem[16,17] and Minimum-cut problem[18]. Among NP-Hard problems, certain instances of the Travelling Salesman Problem have been investigated in [19,20]. A detailed overview of runtime behaviour of Randomized Search Heuristics (RSHs), (1+1) EA and ACO on different pseudo-boolean functions can be found in [21] and on several graph based combinatorial optimization problems can be found in [22].
\par
In this paper, we undertake the runtime analysis of ACO for covering problems on hypergraphs. A hypergraph is a generalization of a graph where an edge, known as hyperedge consists of a non-empty set of vertices with any number of elements. On a hypergraph, different covering problems are the vertex cover and the edge cover problems, both of which can be weighted or unweighted. Independent set problems are also related to covering problems. The decision versions of these problems are NP-complete, while the optimization versions (minimum vertex/edge cover/maximum independent set) are NP-Hard[29]. The vertex cover problem is also known as transversal or hitting set problem, where the edge cover problem is a special case of set cover problem for hypergraphs. Computing the transversal hypergraph has many applications in machine learning, game theory, indexing of databases, SAT problem, data mining and optimization. 
\par
Several interesting results have been obtained by applying various approximation algorithms for determining the vertex cover and set cover problems on a hypergraph or special cases of hypergraphs. Khot has published several results on the approximability and hardness of vertex cover problem on k-uniform hypergraphs[24,26]. In [24], the hypergraph vertex cover problem is shown to have applied directly to several scheduling problems : concurrent open shop and makespan minimization of assembly line. Okun[25] discusses the approximation of hypergraph vertex cover for hypergraphs of bounded degree and bounded number of neighbouring vertices. Complexity and approximation results for the connected vertex cover problem on graphs and hypergraphs [30], vertex cover on dense hypergraphs [27], set covering on hypergraphs [28] and vertex cover on k-partite k-uniform hypergraphs [29] have been obtained. Independent set problems are closely related to covering problems. A comprehensive study of several approximation algorithms on hypergraph independent set problems can be found in [31]. However, the applications of EAs and ACO to these problems are still unexplored.
\par
In the present work, theoretical analysis of ACO is carried out for the minimum weight edge cover problem on generalized hypergraphs. The minimum vertex cover and the maximum weak-independent set problems are shown as special cases of the minimum weight edge cover problem. An incremental construction procedure of ACO is employed in this paper. The expected optimization time is calculated for the MMAS* algorithm with the impact of pheromone values and heuristic information investigated in detail. The expected number of iterations for finding a minimum weight edge cover is found to be of exponential order for generalized hypergraphs without heuristic information present. However, for certain families of hypergraphs, we show that with proper adjustment of the heuristic parameter, expected optimization time of constant order can be achieved. 
\par
The rest of the paper is organized as follows. Section II introduces the fundamental concepts of a hypergraph and gives the problem definitions. Section III gives the detailed description of the MMAS* algorithm and the construction procedure for the minimum weight edge cover problem. In section IV, the upper bound on the expected optimization time is analysed. In section V, the solution methods for the minimum vertex cover and the maximum weak-independent Set problems on hypergraphs are presented as a special case of the minimum weight edge Cover problem. Section VI concludes the paper with a brief discussion on the obtained results.

\section{Preliminaries}
This section gives the definitions to different covering and independent set problems in case of hypergraphs.

\subsection{Basic Definitions:}
\par A \emph{Hypergraph H} is a pair $H$=$(V,E)$, where $V$=\{$v_{1},v_{2},...,v_{n}$\} is a set of discrete elements known as vertices and $E$=\{$e_1,e_2,....e_m$\} is a collection of non-empty subsets of $V$, known as hyperedges. Thus, a hyperedge typically contains any number of vertices. In the rest of the paper, the terms edge and hyperedge will be used interchangeably.
\par The size, or the \emph{cardinality}, \emph{$\left|e\right|$} of a hyperedge is the number of vertices in \emph{e}. A hypergraph is known as \emph{k-uniform} if all the hyperedges have cardinality \emph{k}. A \emph{pendant vertex} is a vertex which is contained in only one hyperedge. 
\par A hypergraph \emph{H(V,E)} is vertex-weighted if every vertex in \emph{V} is assigned a weight. Similarly, a hypergraph is edge-weighted if every edge in \emph{E} is assigned a weight. In this paper, a weighted hypergraph will refer to an edge-weighted hypergraph, unless stated otherwise. 
\par The \emph{dual of a hypergraph}, \emph{H(V,E)} is a hypergraph \emph{H*}, whose vertices and edges are interchanged with \emph{H}. Thus, for each vertex \emph{v} in \emph{H}, there is an hyperedge \emph{e*} in \emph{H*}, formed by vertices in \emph{H*} which correspond to hyperedges incident on \emph{v} in \emph{H}. Thus \emph{e*} = \{$e_{i}\in \emph{E} |\emph{v}\in e_{i}$\}.

\subsection{Covering Problems:}
\par For a hypergraph \emph{H(V,E)}, a \emph{Vertex Cover} or a \emph{Hitting Set} is a subset of \emph{V} that intersects every edge of \emph{H} in at least one vertex. \par Similarly an \emph{Edge Cover} of a hypergraph $H$ is a subset of \emph{E} that contains all vertices in \emph{V}. \textit{Edge Cover} is a special case of the \textit{Set Cover} problem. From the definitions, a hitting set in \emph{H} is a set cover in the dual graph H*. Mathematically, 
\\ \textit{Vertex Cover}: A set \emph{S $\subseteq$ V} such that $\forall$ \emph{$e\in E$}:$\left|e\bigcap S\right|$ $\geq$ 1.
\\ \textit{Edge Cover}: A set \emph{T $\subseteq$ E} such that $\forall$ \emph{$v\in V$}  $\exists$  $e\in T$: \emph{$v \in e$}. 

\par The \textit{Minimum Vertex Cover} and the  \textit{Minimum Edge Cover} of a hypergraph are optimization problems to find the vertex cover and the edge cover of minimum cardinality in a given hypergraph \emph{H}, respectively. Similarly, the  \textit{Minimum Weight Vertex Cover} and \textit{Minimum Weight Edge Cover} problems for edge-weighted and vertex-weighted hypergraphs respectively would mean to find the edge cover and the hitting set of minimum weight. 

\subsection{Independent Set Problems:}
The \emph{Independent Set} problem has two variants in case of hypergraphs. A \emph{Weak-Independent Set} in \emph{H} is a subset of \emph{V} that doesn't contain any edge of H completely. If an independent set intersects any hyperedge in E in at most one element, then it is a \emph{Strong-Independent Set}. Thus for graphs, both the above categories of independent sets are identical. In this paper, independent set refers only to \emph{Weak-Independent Set} unless stated otherwise. Mathematically,    
\\ \textit{Weak-Independent Set}: A set \emph{$I_{w} \subset V$} such that $\forall$ \emph{$e\in E$}:$\left|e\bigcap I_{w}\right|$ $<$ $\left|e\right|$.
\\ \textit{Strong-Independent Set}: A set \emph{$I_{s} \subset V$} such that $\forall$ \emph{$e\in E$}:$\left|e\bigcap I_{s}\right|$ $\leq$ 1.
\par The goal of the \textit{Maximum Weak-Independent Set} of a given hypergraph H is to find a weak-independent set in H of maximum cardinality. Similarly for a vertex-weighted hypergraph H, the \textit{Maximum Weight Weak-Independent Set} problem is to find a weak-independent set in H of maximum weight. 
 
\section{Algorithm}
In this section, we propose an MMAS* (\emph{Max-Min Ant System}) ACO algorithm in the context of \textit{Minimum Weight Edge Cover} problem in an edge-weighted hypergraph. MMAS* is an iterative algorithm, which in each step, constructs a new candidate solution and keeps track of the best-so-far solution. The construction procedure uses a construction graph in which the vertices are the elements to be chosen to generate a new solution. As the name suggests, the pheromone values are bounded between a maximum and minimum value in case of MMAS* algorithm. Thus a typical iteration of MMAS* consists of three steps : \emph{a)Construction of a new solution; b) Selection of the generated solution; c) Pheromone update.} The heuristic information is kept unchanged throughout. 
\subsection{Construction of a new solution:} 
In each iteration, a new solution is obtained by the procedure \textbf {Construct}, as described in Algorithm 1. This is an iterative procedure, in which the artificial ant selects the components of the candidate solution, one at a time, by random walk on the construction graph. For the edge cover problem, the components of the solution are the edges of the input hypergraph \emph{H(V,E)}. Thus, we choose the construction graph \emph{C(H)} to be a directed graph, with vertices \{$e_0,e_1,e_2,....e_m$\}, where $e_i \forall i>0$ corresponds to an edge of \emph{H} and $e_0$ is a starting vertex. The edge set $U$ of C(H) is given as: 
\\  $U$=\{($e_i,e_j$) $\mid$ $0\leq i\leq m, 1\leq j\leq m,i\neq j$\}
\\i.e., in the graph C(H), there is no edge pointing to the starting vertex, $e_0$. 
\par For the edge cover problem, we first select all the edges which contain at least one pendant vertex to be part of the solution by default as per the definition. Then, we make the construction graph omitting all the vertices that correspond to the hyperedges containing pendant vertices in the original hypergraph. In the first step, the artificial ant is at the starting node $e_0$. In each subsequent step, it goes probabilistically to another node in the construction graph avoiding multiple visits to a particular node and only if the new node contains at least one new vertex of the hypergraph which is not covered by the previously visited nodes (i.e. hyperedges of the given hypergraph). Let us assume, without loss of generality, that the $i^{th}$ node traversed is denoted by $e_i$. Hence, we define the notion of feasible neighbourhood \emph{N}($e_k$) for a given node $e_k$ with the previously visited nodes being \{$e_1,e_2,....,e_{k-1},e_k $\} as follows:
\\  $N(e_k):=  E^{\prime} \setminus \{ {e^{\prime}} \in E^{\prime}  \mid  \forall \ v \in e^{\prime}, v \in e_1 \cup e_2 \cup ... \cup e_k \}$, where $E^{\prime}=(E \setminus \{e_1,e_2,....,e_k \})$.
\par Thus, at every step, a new node is selected by the ant within its feasible neighbourhood. The selection process is determined by the current pheromone values $\tau$ of the edges and the heuristic information $\eta$ of the vertices of the construction graph. The probability of selecting any $e_{k+1}$ in the $(k+1)^{th}$ step is given as:     
\\ 
\[
 p_{(e_{k+1})} = \begin{cases}
        \frac{\tau_{(e_k,e_{k+1})}^{\alpha} \cdotp \eta_{e_{k+1}}^{\beta}}{\displaystyle\sum_{e^{\prime} \in N(e_k)} \tau_{(e_k,e^{\prime})}^{\alpha} \cdotp \eta_{e^{\prime}}^{\beta}}, 
           & e_{k+1} \in N(e_k)\\
        0, & otherwise   
        \end{cases}     
 \]
where $\alpha$ and $\beta$ are two positive parameters which control the relative strengths of pheromone value and heuristic information respectively in choosing the new edge of the hypergraph or the node on the construction graph. The construction procedure continues till we get a set of edges such that all the vertices are covered at least once. So, in each step after choosing an edge, we keep track of all the vertices covered so far.  
\par Since, we are interested in the minimum weight edge cover, we would like to prefer edges with less weight and containing more number of vertices. Thus, for a hyperedge $e$, we define the heuristic information as $\eta_e=\frac{\left|e\right|}{w(e)}$. 
\\
\begin{algorithm}
\DontPrintSemicolon
\SetKwFunction{Construct}{Construct}
\SetKwFunction{return}{return}
\Begin{
$V^{\prime}=\phi$
\\$E^{\prime}=\phi$
\\$e \leftarrow e_0$
\\ \While{$V^{\prime} \neq V$}{
Choose a node $e_{k+1} \in N(e_k)$ after choosing $e_k$ with probability
$ \frac{\tau_{(e_k,e_{k+1})}^{\alpha} \cdotp \eta_{e_{k+1}}^{\beta}}{\displaystyle\sum_{e^{\prime} \in N(e_k)} \tau_{(e_k,e^{\prime})}^{\alpha} \cdotp \eta_{e^{\prime}}^{\beta}} $
\\ $E^{\prime} = E^{\prime} \cup e_{k+1}$
\\ $V^{\prime} = V^{\prime} \cup v^{\prime}, \ \forall v^{\prime} \in e_{k+1}$
}
return $E^{\prime}$
}
\caption{The Procedure Construct \label{Construct} on ($C(H),\tau,\eta$)}
\end{algorithm}
\subsection{Fitness Evaluation and Selection:}
The fitness of a candidate solution \emph{x}, $f(x)$=\{$e_1,e_2,...e_k$\} is the total weight of the edges in \emph{x}. Our objective is to minimize this fitness value. Formally, the fitness function is stated as follows:
$f(x):=\displaystyle\sum_{e_i \in x} w(e_i)$.
\par The new solution replaces the best-so-far solution if its fitness value is strictly less than the fitness value of the best-so-far solution.
\subsection{Pheromone update:} 
Initialization of pheromone values is done at the start of the MMAS* algorithm in such a way that their sum is equal to 1. Thus, $\tau_{(u,v)}=1/\left|U\right|; \ \forall(u,v)\in U$.     
\par After an initial solution is \emph{x*} obtained using \textbf{Construct}, the pheromone values are updated using the procedure \textbf {Update}. \textbf {Update}$(\tau,{x^{*}})=\tau^{\prime}$ is defined as follows:
\[
 \forall (e,e^{\prime}) \in U: \tau^{\prime}_{(e,e^{\prime})} = \begin{cases}
        h, & if \ e^{*} \in x^{*}\\
        l, & otherwise   
        \end{cases}     
 \]
where \emph{h} and \emph{l} are the upper and lower bounds on the pheromone values and $h\geq l$. In the subsequent iterations, the procedure \textbf{Update} is used to update the pheromone values, only if the new solution obtained from the procedure \textbf{Construct} replaces the best-so-far solution obtained till the previous iteration. Thus, all the nodes of the construction graph selected in the best-so-far solution have pheromone value \emph{h} on the incoming edges to them. The remaining edges have pheromone value \emph{l}.
\par See Algorithm 2 for the complete algorithm MMAS*.    
\begin{algorithm}
\DontPrintSemicolon
\SetKwFunction{MMAS*}{MMAS*}
\SetKwFunction{Update}{Update}
\Begin {
$\tau_{u,v}\leftarrow 1/\left|U\right|,\forall (u,v)\in U$
\\ $\eta \leftarrow \frac{\left|e\right|}{w(e)}, \forall \ e \in E  $
\\$ x^* \leftarrow Construct(C(H),\tau,\eta)$
\\$\tau \leftarrow Update(\tau,x^*)$
\\ \While{\textit{true}}{
$ x \leftarrow Construct(C(H),\tau,\eta)$
\\ \If{f(x) $<$ f(x*)}{
x*$\leftarrow$ x
}
$\tau \leftarrow Update(\tau,x^*)$
}
}
\caption{The Algorithm MMAS* \label{MMAS*} on \textit{H(V,E)}}
\end{algorithm}

\section{Upper Bounds}
In this section, we show the upper bounds on the expected optimization time of the proposed MMAS* algorithm on the minimum weight edge cover problem of hypergraphs. The results are expressed in terms of the parameters $\alpha$, $\beta$, the pheromone bounds \emph{h} and \emph{l}, the number of hyperedges \emph{m}, and the number of hyperedges in the optimal solution \emph{k} and some other parameters which will be introduced later in this section. First, we do the analysis for generalized hypergraphs and then we show constant order expected optimization time for certain families of hypergraphs.
\subsection{Upper Bounds for the Generalized Hypergraph}
 In case of generalized hypergraphs, we separately investigate the influence of pheromone values and heuristic information respectively.
\newline
\par
\textbf{Theorem 1:} \textit{ Let $\alpha$ $=$ 1 and $\beta$ $=$ 0 and 0  $<$ $l$ $\leq$ $h$, and let c$_{n}$ $=$ $\dfrac{h}{l}$. Then the expected optimization time of the MMAS* is $ O \left( \frac{((m-k)c_{n} + k)!}{((m-k)c_{n})!k!}\right) $.}
\\

\begin{proof} For the purpose of calculating the upper bound, we first need to find a lower bound on the probability that the MMAS* algorithm gives the optimal solution. We consider the worst possible case . Thus, we  assume that there are no edges having vertices with degree 1, the \emph{k} edges belonging to the optimal solution have pheromone level \emph{l} and the rest of the edges have pheromone level \emph{h}. The probability of selecting an edge belonging to the optimal solution in step \emph{i} is at least:

\begin{align}
\nonumber \frac{(k-i)l}{(m-k-i)h+(k-i)l}  = \frac{(k-i)l}{((m-k)h+kl)-i(h+l)} \\
 \nonumber = \frac{k-i}{((m-k)c_{n}+k)-i(c_{n}+1)}\\
\nonumber \end{align}

The probability that the MMAS* algorithm gives the optimal solution is at least:
\[  \displaystyle\prod_{i=0}^{k-1} \frac{k-i}{((m-k)c_{n}+k)-i(c_{n}+1)} = \frac{k! ((m-k)c_{n})!}{((m-k)c_{n}+k)!}\]

Hence, the expected optimization time is found to be of \[ O \left( \frac{((m-k)c_{n}+k)!}{k! ((m-k)c_{n})!} \right) \]
\end{proof}
Now, we move to the second scenario, where we analyse the upper bound for the expected optimization time  when $\alpha = 0$ and $\beta = 1$. Here we examine the behaviour of the algorithm in the presence of heuristic information. \\ 

\textbf{Theorem 2:} \textit{ Choosing $\alpha$ $=$ 0 and $\beta$ $=$ 1, the expected optimization time of the MMAS* is bounded by $ O \left[\left( \frac{\eta_{max}}{\eta_{min}}\right) (m-k) \right] ^k$. }
\\

\begin{proof}Let $e_{1},e_{2},......,e_{m}$ be the hyperedges of the hypergraph and $\eta_{1},\eta_{2},......,\eta_{m}$ be their corresponding heuristic information values. Let $T = \left\lbrace \eta_{1}, \eta_{2},....., \eta_{k},.......,\eta_{m} \right\rbrace $denote the set that contains the heuristic information of all the hyperedges.  Without loss of generality, let $S = \left\lbrace \eta_{1}, \eta_{2},....., \eta_{k} \right\rbrace $ be the set that contains the heuristic information of the edges present in the optimal solution. Let $P_{i}$ denote the probability that the $i^{th}$ hyperedge selected during the course of one iteration, belongs to S. Also, let $R_{i}$ be the set containing the elements of S which have not been included in the solution till the $(i-1) th$ iteration.

\[ P_{i} = \frac{\displaystyle\sum_{\eta_{j} \in R_{i}} \eta_{j}^{\beta}}{\displaystyle\sum_{\eta_{j} \in ((T \setminus S) \cup R_{i}} \eta_{j}^{\beta}} 
= \frac{\displaystyle\sum_{\eta_{j}\in R_{i}} (\eta_{min}^{\beta} + C_{j})}{\displaystyle\sum_{\eta_{j}\in R_{i}} (\eta_{min}^{\beta}  + C_{j}) + \displaystyle\sum_{\eta_{j}\in (T \setminus S)} \eta_{j}^{\beta}}  \]
\[\geq \frac{\displaystyle\sum_{\eta_{j}\in R_{i}} \eta_{min}^{\beta}}{\displaystyle\sum_{\eta_{j}\in R_{i}} \eta_{min}^{\beta} + \displaystyle\sum_{\eta_{j}\in (T \setminus S)} \eta_{j}^{\beta}} 
\geq \frac{\displaystyle\sum_{\eta_{j}\in R_{i}} \eta_{min}^{\beta}}{\displaystyle\sum_{\eta_{j}\in R_{i}} \eta_{min}^{\beta} + \displaystyle\sum_{\eta_{j}\in (T \setminus S)} \eta_{max}^{\beta}} \]
\[ \geq  \frac{\eta_{min}^{\beta}}{\eta_{min}^{\beta} + (m-k)\eta_{max}^{\beta}}
= \frac{1}{1 + \left(\frac{\eta_{max}}{\eta_{min}}\right)^{\beta} (m-k)}\]
where $\eta_{min}=min \left\lbrace \eta \mid \eta \in T \right\rbrace$ , $\eta_{max}=max \left\lbrace \eta \mid \eta \in T \right\rbrace$ and $C_{j} = \eta_{j} - \eta_{min} $ for $ \eta_{j} \in S\ ,\  j=1 $ to $ k$, are positive constants.
Therefore,  $P_{MIN}$, the lower bound on the probability of choosing $S$ as the solution is:
\[ P_{MIN} = \frac{1}{ \left[ 1 + \left(\frac{\eta_{max}}{\eta_{min}}\right)^{\beta} (m-k) \right]^{k}} \]
Substituting  $\beta = 1$ and taking the inverse of $P_{MIN}$, we find the expected optimization time to be of    
 \[ O \left[ 1 + \left( \frac{\eta_{max}}{\eta_{min}}\right) (m-k) \right]^{k}  i.e. \  O \left[\left( \frac{\eta_{max}}{\eta_{min}}\right) (m-k) \right]^{k} \].\end{proof}

It can thus be noted that the expected optimization time of the MMAS* algorithm depends on the nature of the graph. In the worst case though, the parameter $k$ stated above, has an upper bound of $n$. It can be shown that there exists a certain family of graphs for which the expected optimization time is of constant order.

\subsection{Upper Bounds for Specific Instances of Hypergraphs}

We consider only the case where $\alpha = 0$ and $\beta = 1$ since this case yields better results as compared to the case where $\alpha = 1$ and $\beta = 0$. It may thus be noted that heuristic information is more important than pheromones for faster convergence. \\

\textit{Definitions:} Before proceeding to theorem 3, let us define the terms $\eta_{1 max}$ and $\eta\prime_{min}$.
\[\eta_{1 max} = max \left\lbrace \eta \mid \eta \in T \setminus S \right\rbrace \] \[ \eta\prime_{min} = min \left\lbrace \eta \mid \eta \in S \right\rbrace \]
where $S$ and $T$ are as defined in the proof of theorem 2.\\

\textbf{Theorem 3:} \textit{ If $ \eta \prime_{min} > \eta_{1 max}$ and we choose $ \beta \geq \frac{log(k(m-k))}{log \left( \eta \prime_{min} / \eta_{1 max} \right) }$, the expected optimization time for the MMAS* algorithm is of constant order.}
\\

\begin{proof} Using the definitions mentioned in the proof of Theorem 2 and proceeding in a similar fashion, we obtain:
\[ P_{i} = \frac{\displaystyle\sum_{\eta_{j} \in R_{i}} \eta_{j}^{\beta}}{\displaystyle\sum_{\eta_{j} \in ((T \setminus S) \cup R_{i}} \eta_{j}^{\beta}} 
= \frac{\displaystyle\sum_{\eta_{j}\in R_{i}} (\eta\prime_{min}^{\beta} + C^{\prime}_{j})}{\displaystyle\sum_{\eta_{j}\in R_{i}} (\eta\prime_{min}^{\beta}  + C^{\prime}_{j}) + \displaystyle\sum_{\eta_{j}\in (T \setminus S)} \eta_{j}^{\beta}}  \]
\[\geq \frac{\displaystyle\sum_{\eta_{j}\in R_{i}} \eta\prime_{min}^{\beta}}{\displaystyle\sum_{\eta_{j}\in R_{i}} \eta\prime_{min}^{\beta} + \displaystyle\sum_{\eta_{j}\in (T \setminus S)} \eta_{j}^{\beta}} 
\geq \frac{\displaystyle\sum_{\eta_{j}\in R_{i}} \eta\prime_{min}^{\beta}}{\displaystyle\sum_{\eta_{j}\in R_{i}} \eta\prime_{min}^{\beta} + \displaystyle\sum_{\eta_{j}\in (T \setminus S)} \eta_{1 max}^{\beta}} \]
\[ \geq  \frac{\eta\prime_{min}^{\beta}}{\eta\prime_{min}^{\beta} + (m-k)\eta_{1 max}^{\beta}}
= \frac{1}{1 + \left(\frac{\eta_{1 max}}{\eta\prime_{min}}\right)^{\beta} (m-k)}\]

where  $C^{\prime}_{j} = \eta_{j} - \eta^{\prime}_{min} $ for $\eta_{j} \in S\ ,\   j=1$ to $k$ are positive constants. \\

Therefore, $P^{\prime}_{MIN}$, the lower bound on the probability of choosing $S$ as the solution is:
\[ P^{\prime}_{MIN} = \frac{1}{ \left[ 1 + \left(\frac{\eta_{1 max}}{\eta\prime_{min}}\right)^{\beta} (m-k) \right]^{k}} \]

In the expression of $P^{\prime}_{MIN}$, if we choose $\left[ \left(\frac{\eta_{1 max}}{\eta\prime_{min}}\right)^{\beta} (m-k)\right]$ to be bounded above by $\frac{1}{k}$, we would obtain:
\[  P^{\prime}_{MIN}\ \geq \left(1+\frac{1}{k}\right)^{-k} \geq \frac{1}{e} \]

Then, the expected optimization time of MMAS* would be bounded above by the inverse of $P^{\prime}_{MIN}$ i.e. of $O(e)$, which is of constant order.\\

For the above assertion to be true, the condition $\left[ \left(\frac{\eta_{1 max}}{\eta\prime_{min}}\right)^{\beta} (m-k)\right] \leq \frac{1}{k}$ must hold for some non-negative $\beta$. Two cases may arise.\\

\textit{Case 1:} $ \left(\frac{\eta_{1 max}}{\eta\prime_{min}}\right) \geq 1$\\

This case will never hold good for any non-negative $\beta$, and hence is rejected.\\

\textit{Case 2:} $ \left(\frac{\eta_{1 max}}{\eta\prime_{min}}\right) < 1$\\

\[\left[ \left(\frac{\eta_{1 max}}{\eta\prime_{min}}\right)^{\beta} (m-k)\right] \leq \frac{1}{k}\]
\[ \Rightarrow \beta \ log\left(\frac{\eta_{1 max}}{\eta\prime_{min}} \right) \leq log \left( \frac{1}{k(m-k)}\right) \]
\[ \Rightarrow \beta \geq \frac{log(k(m-k))}{log \left( \frac{\eta\prime_{min}}{\eta_{1 max}} \right)}\]\\

Hence, it can be concluded that, if $ \eta \prime_{min} > \eta_{1 max}$ and we choose $ \beta \geq \frac{log(k(m-k))}{log \left( \eta \prime_{min} / \eta_{1 max} \right) }$, the expected optimization time for the MMAS* algorithm is of constant order.\\ \end{proof}

Some instances of hypergraphs for which the above conditions hold good and thus the MMAS* algorithm has constant order expected optimization time, are discussed below.\\

\textit{Instance 1:} We shall construct a weighted complete $r$-uniform hypergraph $H_1 = (V_1,E_1)$ on the vertex set $V_1= \left\lbrace 1,2,.....,n \right\rbrace$ and the hyperedge set $E_1= \left\lbrace e_1,e_2,.....,e_m \right\rbrace$, where each hyperedge $e$ is treated as a set of vertices. Let $S=\phi$ and $T=\phi$ be empty sets and $i$ be initialized to $0$. Let weight of an edge $e$ be denoted by $w(e)$. We construct the hypergraph as follows:\\
$(i)$ increment $i$ and select a new hyperedge $e^\prime_i \in (E \setminus S)$ such that $ T \cap e^\prime_i = \phi$\\ 
$(ii)$ assign $w(e^\prime_i)=1$. Update $S=S\cup \left\lbrace e^\prime_i \right\rbrace$ and $T=T \cup  e^\prime_i $\\
$(iii)$ if $i < \lfloor \frac{n}{r} \rfloor $, repeat steps $(i)$, $(ii)$ and $(iii)$. Else, goto step $(iv)$\\
$(iv)$ increment $i$ and select a new hyperedge $e^{\prime}_{i}$ such that $( V \setminus T) \subseteq e^{\prime}_{i}$\\
$(v)$ assign $w(e^\prime_i)=1$. Update $S=S\cup \left\lbrace e^\prime_i \right\rbrace$ and $T=T \cup  e^\prime_i $\\
$(vi)$ set $w(e_j) = rand_j \   \forall \ e \in (E \setminus S)$, where $rand_j$ is a random integer $> 1 $\\

Clearly, for the above hypergraph $H_1$, the minimum weight edge cover is the set $S_1$. Also, since H satisfies the conditions of Theorem 3, expected optimization time would be of constant order.
\newline 
\par 
\textit{Instance 2:} We construct an unweighted hypergraph $H_2 = (V_2,E_2)$ on the vertex set $V_2= \left\lbrace 1,2,.....,n \right\rbrace$ and the hyperedge set $E_2= \left\lbrace e_1,e_2,.....,e_m \right\rbrace$. Let $S=\phi$ and $T=\phi$ be empty sets and $i$ be initialized to $0$. Let $p$ be an integer such that $2 \leq p \leq n$.\\
$(i)$ increment $i$ and construct a new hyperedge $e^\prime_i$ such that $e^\prime_i \notin S$ as follows:\\
  $\quad \vert e^\prime_i \vert = p$\\
  $\quad $ if $p \leq \vert V \setminus T \vert $, then $e^{\prime}_{i} \subseteq (V \setminus T)$\\
  $\quad $else $ (V \setminus T) \subset e^{\prime}_{i}$\\
$(ii)$ update $p=q$ such that $ 2\leq q\leq p$, $S=S\cup \left\lbrace e_i \right\rbrace$ and $T=T \cup  e_i $\\
$(iii)$ if $\vert T \vert < \vert V \vert$, then repeat steps $(i)$ and $(ii)$, else goto step $(iv)$.\\
$(iv)$ terminate construction procedure if so desired or if the maximum limit for the number of hyperedges i.e. $2^n$ is reached.
$(v)$ update $p=q$ such that $ 2\leq q\leq p$ and $S=S\cup \left\lbrace e_i \right\rbrace$ .\\
$(vi)$ increment $i$ and construct a new hyperedge $e^\prime_i$ such that $e^\prime_i \notin S$ and $ \vert e_i \vert = p$.\\

The expected optimization time of the MMAS* algorithm for the above mentioned \textit{Instance 2} of hypergraphs, is of constant order.\\ 

\begin{proof} Let the set of all hyperedges belonging to $S$ that were constructed using step $(i)$ be denoted as $\Psi$. Let $\Phi = S \setminus \Psi$. Let the last hyperedge that was constructed using step $(i)$ be denoted by $e_{last}$. Since the hypergraph is unweighted, we may assume that each hyperedge has weight equal to 1. Now, in order to show that $\Psi$ is the minimum edge cover of the hypergraph $H_2$, we need to show that no subset of $\Psi$ can be replaced by any subset of  $\Phi$ while not increasing the total number of edges used in the edge cover. Let $U \subseteq (\Psi \setminus \{ e_{last} \})$ and $V \subseteq \Phi$. For $V$ to be able to replace $U$, there must exist some $V$ s.t. $\vert V \vert < \vert U \vert $ and $V$ covers all vertices covered by $U$. However if $\vert V \vert < \vert U \vert $, the number of vertices covered by $V$ is less than the number of vertices covered by $U$ since $\forall  (e_U , e_V) \ s.t. \  e_U \in U$ and $e_V \in V, \  \vert e_U \vert \geq \vert e_V \vert$ and elements of $U$ are disjoint while elements of $V$ are not necessarily disjoint. \\

Hence, for no $U$ can a corresponding $V$ exist, which can replace $U$ while satisfying the condition $\vert V \vert < \vert U \vert $. \hfill (1)\\

Also, for no $(U \cup \{e_{last}\})$ can a corresponding $V$ exist, which can replace $(U \cup \{e_{last}\})$ while satisfying the condition $\vert V \vert \leq \vert U \vert $. \hfill (2)\\

 Hence, using statements (1) and (2), it can be concluded that, no subset of $\Psi$ can be replaced by any subset of  $\Phi$ while not increasing the total number of edges used in the edge cover.  Thus, $\Psi$ is a minimum vertex cover. 
\par Applying theorem 3, there exists $\beta$ s.t. the expected optimization time for the MMAS* algorithm is of constant order.

\end{proof}

\section{Other Hypergraph covering problems}

\subsection{Minimum Vertex Cover Problem}
The vertex cover, as defined earlier, is a set of vertices such that each hyperedge contains at least one vertex belonging to this set. The minimum vertex cover problem is finding a vertex cover of minimum cardinality. 

The minimum edge cover problem is a special case of the set cover problem where we treat each set as a hyperedge and the union of all the given sets as the set of vertices. Similarly, the minimum vertex cover problem is a special case of the hitting set problem. It is known that the set cover problem and the hitting set problem are equivalent reformulations of one another. Thus, for hypergraphs,  the minimum vertex cover problem is an equivalent reformulation of the minimum edge cover problem. 

Hence, we can address the minimum vertex cover problem by reducing to the minimum edge cover problem, for which we have already provided an MMAS* algorithm. For this purpose, we take the given hypergraph $H(V,E)$ as input, find its dual hypergraph $H^*(V^*,E^*)$ with unweighted edges and then apply the MMAS* algorithm for finding the minimum edge cover to $H^*$. The MMAS* algorithm gives a set of hyperedges as output which is a subset of $E^*$. By the definition of dual of a hypergraph, these set of hyperedges  of $H^*$ given as output, are vertices of the hypergraph $H$ and hence form a solution for the minimum vertex cover problem for the hypergraph $H$.

\subsection{Maximum-Weak Independent Set Problem}
We know that a given set of vertices forms a weak-independent set if and only if its complement is a vertex cover. Hence, the solution to the weak independent set problem can be obtained directly from the solution of the minimum vertex cover problem, by taking the complement of the minimum vertex cover.

It should also be noted that the maximum clique problem can solved by taking a complement hypergraph $H^\prime$ of the input hypergraph $H$ and then finding the maximum weak-independent set for $H^\prime$.

\section{Conclusion}
In the present work, we obtain results for the upper bound of expected optimization time of MMAS* ACO algorithm for the minimum weight edge cover problem on generalized hypergraphs by taking into account pheromone values and heuristic information separately. The results obtained are of exponential order with respect to the number of hyperedges for generalized hypergraphs. However, for certain families of hypergraphs, by suitably choosing the heuristic parameter, expected optimization time of constant order could also be obtained. We show two such theoretical instances where the above holds true. Also, we show that the minimum vertex cover and maximum weak-independent set problems are special cases of the minimum weight edge cover problem. 
\par Further work may be taken up in the following directions.
\begin{itemize}
    \item Investigation of different ACO construction procedures and corresponding runtime analysis for the covering problems on specific classes of hypergraphs, i.e. k-uniform, k-partite, dense/sparse hypergraphs, etc.
    \item In-depth analysis of the impact of pheromone values and the pheromone evaporation factor on the running time.
    \item Analysis of lower bounds on the expected optimization time for different classes of hypergraphs.       
    \item Extension of the present work to other NP-Hard/NP-Complete problems on hypergraphs, eg. coloring problems, minimum spanning tree problem, etc.
\end{itemize}  
\par The present study thus points to several interesting open areas in the theoretical ACO research that may contribute to a better understanding of the runtime behaviour of ACO for computationally harder problems.  

\section{Acknowledgement}
The author thanks Ashish Ranjan Hota for initial discussions which provided him with valuable insight into the problem. 


%

\def\V{\rm vol.~}
\def\N{no.~}
\def\pp{pp.~}
\def\Pot{\it Proc. }
\def\IJCNN{\it International Joint Conference on Neural Networks\rm }
\def\ACC{\it American Control Conference\rm }
\def\SMC{\it IEEE Trans. Systems\rm , \it Man\rm , and \it Cybernetics\rm }

\def\handb{ \it Handbook of Intelligent Control: Neural\rm , \it
    Fuzzy\rm , \it and Adaptive Approaches \rm }

\end{document}